\documentclass[runningheads]{llncs}
\usepackage[T1]{fontenc}
\usepackage{graphicx}
\usepackage[english]{babel}
\usepackage{hyphenat}
\usepackage{booktabs}
\usepackage{amssymb}

\usepackage{float}

\usepackage[verbose]{hyperref}

\begin{document}
%
\title{Scalable AI Inference: Performance Analysis and Optimization of AI Model Serving}

\titlerunning{Pham and Gedikli}
%
\author{Hung Cuong Pham\inst{1} \and Fatih Gedikli\inst{1}\orcidID{0000-0001-6190-0449  }}
\authorrunning{Pham and Gedikli}
%
\institute{Institute of Computer Science \\
University of Applied Sciences Ruhr West \\
Mülheim an der Ruhr, Germany \\
\email{hung.pham@stud.hs-ruhrwest.de, fatih.gedikli@hs-ruhrwest.de}
}
\maketitle              
\begin{abstract}
AI research often emphasizes model design and algorithmic performance, while deployment and inference remain comparatively underexplored despite being critical for real-world use. This study addresses that gap by investigating the performance and optimization of a BentoML-based AI inference system for scalable model serving developed in collaboration with graphworks.ai. The evaluation first establishes baseline performance under three realistic workload scenarios. To ensure a fair and reproducible assessment, a pre-trained RoBERTa sentiment analysis model is used throughout the experiments. The system is subjected to traffic patterns following gamma and exponential distributions in order to emulate real-world usage conditions, including steady, bursty, and high-intensity workloads. Key performance metrics, such as latency percentiles and throughput, are collected and analyzed to identify bottlenecks in the inference pipeline. Based on the baseline results, optimization strategies are introduced at multiple levels of the serving stack to improve efficiency and scalability. The optimized system is then reevaluated under the same workload conditions, and the results are compared with the baseline using statistical analysis to quantify the impact of the applied improvements. The findings demonstrate practical strategies for achieving efficient and scalable AI inference with BentoML. The study examines how latency and throughput scale under varying workloads, how optimizations at the runtime, service, and deployment levels affect response time, and how deployment in a single-node K3s cluster influences resilience during disruptions.
\end{abstract}

\section{Introduction}

In machine learning, applying a previously trained model to new data in order to generate predictions is referred to as the inference phase. Unlike training, which is typically performed offline and focuses on improving model quality, inference takes place in user-facing environments where response time, throughput, and service availability directly affect the usability of the system. End users interact with machine learning services during this phase to receive predictions, recommendations, or insights based on their input. As a result, efficient and scalable model serving has become a critical requirement for deploying ML systems in production.

Despite the rapid progress in machine learning research, most attention remains focused on model architectures, training methods, and benchmark accuracy, while comparatively less emphasis is placed on deployment and inference. However, strong predictive performance alone is not sufficient in practice if a model cannot be served reliably, efficiently, and at scale. This creates a gap between model-centric research and the operational demands of real-world AI systems.

Motivated by this gap, this study evaluates the performance and optimization of a BentoML-based predictor for scalable AI inference. The system is analyzed with respect to inference latency, throughput, and resilience under different workload patterns and deployment configurations. By examining optimizations at the model, service, and deployment levels, this work aims to provide practical insights into building an efficient, scalable, and resilient AI inference service for production environments.

\section{Related Work}
\subsection{Model Serving}
While machine learning research has traditionally focused on model architectures, training methods, and predictive performance, a growing body of work has begun to address the systems challenges of deploying models in production. Model serving has emerged as an important area of research concerned with how trained models are exposed as reliable, efficient, and scalable services. This includes challenges such as minimizing inference latency, maximizing throughput, handling fluctuating workloads, and maintaining service resilience under operational disruptions.

Much of the recent literature on model serving has focused on large language models (LLMs), whose high computational demands make efficient inference particularly challenging. For example, Miao et al. \cite{Miao_2024} introduced SpecInfer, which accelerates the serving of generative LLMs through tree-structured speculative inference and verification. Their approach reduces end-to-end latency while preserving model quality.

Strati et al. \cite{10.5555/3692070.3693972} proposed Déjà Vu, a solution that reduces internal costs associated with distributed LLM serving. It uses state replication for fault tolerance, micro-batch switching for efficient GPU memory management, and prompt-token disaggregation to minimize pipeline bubbles.

Wang et al. \cite{10.1145/3711896.3737413} presented the first real-world trace dataset focused on LLM serving workloads, including user, system, and LLM actions. The authors also created a benchmark suite based on these traces to facilitate the performance assessment of various serving systems.

Miao et al. \cite{10.1145/3620665.3640411} proposed a solution focusing on implementing LLM serving systems on preemptive instances to balance cost and performance. 

Duan et al. \cite{10.5555/3692070.3692543} argued that efficiently serving multiple large language models is challenging because of their diverse usage patterns. To address this issue, Duan et al. introduced MaxServe, a system that improves efficiency by strategically colocating models based on their popularity and separating the prefill and decoding phases. This optimizes memory usage and computing resources. Their solution includes algorithms for optimal placement and adaptive scheduling. As a result, MaxServe boosted performance significantly, achieving up to 1.8 times higher throughput and handling 2.9 times more requests while maintaining reliability.

\subsection{Benchmarking}
Numerous ML inference benchmarks have been created by academic and industrial organizations. Examples from industry include AIMatrix\footnote{https://aimatrix.ai/en-us/} , AIXPRT\footnote{https://www.principledtechnologies.com/benchmarkxprt/aixprt/}, and EEMBC MLMark\footnote{https://www.eembc.org/mlmark/}. Academic organizations have also developed many benchmarks, such as TBD \cite{8573476}, Fathom \cite{Adolf_2016}, and DAWNBench\footnote{https://dawn.cs.stanford.edu/dawnbench/}.

In a research paper, Reddi et al. \cite{9138989} presented MLPerf Inference, a standard machine learning inference benchmark suite. The suite provides proper metrics and a benchmarking method that allow for fair measurement of the inference performance of ML hardware, software, and services. Through their research, Reddi et al. explained how standardizing evaluation techniques, developing realistic ML inference situations, and constructing appropriate ML benchmarking metrics make realistic performance optimization for inference quality possible.

Rodriguez et al. \cite{10.1145/3441250.3441277} analyzed the benchmark results of two well-known benchmarks, MLMark and MLPerf. The researchers aimed to provide a basis for comparing the two benchmarks and to offer recommendations on computer architectures for utilizing inference in machine learning (ML).

In their research, Chitty-Venkata et al. \cite{10.1109/SCW63240.2024.00178} introduced the LLM-Inference-Bench, a benchmarking suite designed to evaluate the performance of large language models on different types of hardware. The study revealed that LLMs require substantial computational resources. This requirement made hardware efficiency more important. In their research, they compared various models and frameworks, as well as GPUs and specialized AI accelerators, and revealed their strengths and weaknesses. Additionally, their solution provided an interactive dashboard that allowed users to select the optimal hardware and configuration for their needs.

In a paper, Rosa et al. \cite{10749741} evaluated the performance of five widely used model serving frameworks: These frameworks are TensorFlow Serving, TorchServe, MLServer, MLflow and BentoML. The authors analyzed their performance by subjecting them to four different scenarios: malware detection, cryptocurrency price forecasting, image classification, and sentiment analysis.

Sonia et al. \cite{10.1145/3533028.3533308} conducted the first analysis of the performance of model serving integration tools in stream processing frameworks. They discovered that pipelines feeding pre-trained models with embedded libraries had higher throughput. Additionally, TensorFlow Serving outperformed TorchServe consistently.

\subsection{Load Testing}
Load testing is a crucial software quality assurance practice that evaluates a system’s capacity to function under expected user demand. Jiang and Hassan \cite{7123673} conducted a comprehensive review of research and practices related to designing, executing, and analyzing load tests for large-scale systems. They assumed that load testing could serve multiple functional and non-functional purposes, each with its own pass/fail criteria. This differs from functional testing, where the objective is clear.

Jiang et al. \cite{harrold2000testing} emphasized the importance of creating realistic workloads. They did so by formulating research questions on load testing based on the three phases of traditional software testing: test design, execution, and analysis. They also highlighted the difference between synthetic and field workloads.

Similarly, Pargaonkar \cite{articleShravan} provided a thorough review of performance testing methodologies. He focused on technologies such as cloud-based platforms, containerization, continuous testing pipelines, and performance metric selection. The author also discussed the challenges of workload modeling and real-world environmental simulation in his review.

A recent study by Dumitrescu et al. \cite{Dumitrescu} concluded that a scalable testing infrastructure is essential for evaluating high-traffic systems. The researchers developed an automated framework that coordinates distributed load agents to test service throughput and fairness.

\subsection{Container Management and Orchestration}
A container is a lightweight, portable unit of software that packages an application together with its runtime environment, dependencies, and configuration, allowing it to run consistently across different computing environments. Containers are widely used in modern software deployment because they provide process isolation while sharing the host operating system kernel, making them more efficient than traditional virtual machines.

An optimal container management process minimizes the effort required to create, delete, deploy, configure, and scale a large number of containers on a host. It also automates the deployment of container-based applications across different infrastructures, thereby increasing interoperability \cite{7036275,7922500,8861307}. A centralized orchestration tool is typically required for managing multiple containers. These tools can be described as frameworks that integrate, manage, control, and define containers on a large scale \cite{8944745}. Additionally, they enable automated workflows for deployment and change management \cite{8762053}. According to a recent survey \cite{8944745}, the adoption of these tools has doubled every year since 2015. 

\subsection{Research Gaps}



Modern AI applications must operate under strict latency, throughput, and reliability constraints, particularly when serving large models or handling variable workloads. Consequently, inference-system metrics such as latency, throughput, and resilience to disruptions are critical to overall system effectiveness. Although many tools and frameworks exist for model serving, their behavior under realistic traffic patterns and deployment configurations is still not sufficiently understood.

This study focuses on BentoML\footnote{https://www.bentoml.com/}, a model-serving framework chosen because it is specifically designed to simplify the packaging, deployment, and management of ML models as production services. Its support for service definition, model integration, and deployment workflows makes it a relevant platform for studying practical inference optimization. However, performance evaluations and optimization analyses of BentoML-based predictors, especially under stochastic workloads and in container-orchestrated environments, remain limited.

To examine deployment-level behavior, the study also uses K3s\footnote{https://k3s.io/}, a lightweight Kubernetes distribution selected for its low operational overhead and suitability for resource-constrained or simplified cluster environments. K3s provides a practical orchestration platform for investigating how deployment mechanisms affect the resilience of BentoML-based inference services during failure scenarios. Despite its increasing adoption, its role in supporting robust AI inference under disruption is not yet well documented.

Therefore, it is necessary to empirically evaluate the behavior of a BentoML inference service under different optimization strategies, workload conditions, and deployment configurations, while also examining how a K3s-based environment influences service resilience.

\section{Approach}

\subsection{Requirements}
This section outlines the functional and non-functional requirements of the proposed system. From a functional perspective, the system must be able to retrieve machine learning models from a registry and make them available for inference requests. It must also support scalable serving to handle increases in incoming traffic, for example through orchestration platforms such as Kubernetes.

From a non-functional perspective, the system must provide low-latency responses to inference requests in order to remain suitable for real-time or interactive use. In addition, it should scale efficiently as the number of users or requests grows. The service must also be resilient, meaning that it should recover automatically from disruptions with minimal downtime and without requiring manual intervention.

\subsection{Study Design}
The study is structured into the following main stages:
\begin{enumerate}
    \item \textbf{Replace existing simpletransformers\footnote{https://simpletransformers.ai/} implementation with Hugging Face Transformers and ONNX Runtime:} By adopting Hugging Face Transformers for PyTorch models and ONNX Runtime for ONNX models, users gain full control and flexibility. They have fine-grained access to the model's architecture, tokenization, and training loop. This makes customization for research or advanced use cases easier. In contrast, the simpletransformers abstraction hides many important details and limits control over training and inference. Consequently, debugging and customization are more difficult, and it is not ideal for large-scale or production systems.
    \item \textbf{Perform Optimizations at the Model, Runtime, Service, and Deployment Levels:}
    \begin{itemize}
        \item \textbf{Optimizations at Model Level:} First, the base FP32 PyTorch model is converted into an FP32 ONNX model. Then, this ONNX model is improved by optimizing its graph structure for faster execution and reduced memory usage without changing its output. Finally, the optimized model is converted into FP16 format. Additionally, the performance of inference in FP16 with PyTorch is evaluated. Rather than converting to FP16 manually, Hugging Face Transformers enables users to load the PyTorch model with all weights directly in FP16 instead of FP32. 
        \item \textbf{Optimizations at Runtime Level:} For the inference with PyTorch in Hugging Face Transformers, the gradient tracking and dropout, batchnorm training behavior are disabled.
        \item \textbf{Optimizations at Service Level:} Enabling adaptive batching in BentoML allows multiple requests to be grouped dynamically for more efficient processing. Depending on real-time traffic, the batch size (the maximum number of requests per batch) and the batch window (the maximum amount of time the service waits before grouping requests into a batch) continuously adjust.
        \item \textbf{Optimizations at Deployment Level:} In this study, the service is deployed to a K3s cluster, a lightweight Kubernetes distribution. This cluster consists of a single node with one replica. K3s can provide automated recovery and self-healing behavior for the service with only one replica and a single node, which is not possible with the current Docker-only deployment.
    \end{itemize}
    \item \textbf{Perform Model Evaluation:} The unoptimized implementation and the optimized implementations are evaluated across batch sizes ranging from 1 to 32 with identical datasets. The unoptimized baseline implementation with simple transformers and FP32 PyTorch serves as a reference. The optimized implementations use either Hugging Face Transformers with PyTorch models or the ONNX Runtime with ONNX models. 
    \item \textbf{Perform Load Testing:} Each evaluation type undergoes load testing under three different traffic scenarios. These scenarios are scaled to maintain a constant mean inter-arrival time of 2 seconds, ensuring that the only difference across traffic patterns is the variance (burstiness). 
        \begin{itemize}
            \item \textbf{Steady Traffic:} The traffic has a constant average arrival rate of $\lambda$ = 0.5 and is modeled as a Poisson process. Therefore, the inter-arrival times follow an exponential distribution with a mean of two seconds.
            \item \textbf{Traffic with Moderate Burstiness:} This pattern is modeled using a Gamma distribution with a shape parameter $\alpha$ = 1.2. Compared to the exponential case ($\alpha$ = 1), this results in slightly reduced variability and more regular inter-arrival times, all while maintaining the same mean.
            \item \textbf{Traffic with Extreme Burstiness:} This pattern is modeled using a Gamma distribution with a shape parameter $\alpha$ = 0.8. In this case, a smaller $\alpha$ value leads to increased variance, resulting in highly bursty traffic characterized by clusters of arrivals and longer inter-arrival intervals.
        \end{itemize}
    \item \textbf{Perform Resilience Testing:} During the resilience test, the service endpoint will undergo continuous load testing. Conducting load testing during a disruption simulation demonstrates the system’s ability to handle traffic during failures and its resilience. The system will continue servicing traffic with minimal downtime as Kubernetes automatically detects and resolves failures. This is the objective of the experiment. However, due to deployment to a single-node cluster with only one replica, the simulated disruptions are limited. The disruptions simulated in this study are: 
    \begin{itemize}
        \item Disruption of the Kubernetes Control Plane
        \item Termination of Containerd, the default built-in container runtime for K3s
        \item Restarting a Deployment via Rollout
        \item Termination of Kubernetes ReplicaSets
        \item Termination of Kubernetes Pods
    \end{itemize}
    \item \textbf{Results and Discussion:} The final results are shown. These results reveal the effectiveness of the optimization approaches on the model performance, load testing, and resilience testing. The findings are analyzed in detail, along with a discussion of the practical implications. Furthermore, future directions are presented to improve the system's inference performance.
\end{enumerate}
\subsection{Traffic Modeling}
In the context of system performance and networking, "burstiness" refers to traffic that arrives in brief, intense bursts, followed by quieter intervals, rather than arriving steadily and uniformly. Burstiness is an important concept in this work because AI model serving systems must often handle unpredictable user behavior in real-life scenarios. High burstiness can overwhelm queuing systems, autoscalers, and model load times. This study will simulate burstiness during load testing to examine the responses of the BentoML approach and the custom stack to traffic spikes.

Load testing tools such as JMeter and Locust have their limitations. Their ramp-up does not accurately reflect real-world traffic. In reality, users arrive randomly rather than at predictable, constant rates. Additionally, fixed-rate injection, in which users are added at a constant rate, can result in sharp spikes in system load due to sudden increases in traffic. Consequently, the system may appear unstable even though it can perform well. To make traffic load testing more realistic, Poisson and gamma distributions are used to simulate request arrival patterns. In their research, Wang et al. \cite{10.1145/3711896.3737413} used the Gamma distribution to simulate burstiness when benchmarking LLM serving systems.

Unlike load generation based on a fixed user spawn rate, stochastic inter-arrival models introduce probabilistic delays between requests. Consequently, the throughput achieved during the experiment differs from the nominal request rate, which is generally lower than the rate obtained using the deterministic spawning provided by the load test tool. Therefore, the requests per second (RPS) values reported by Locust in this study reflect the system’s throughput under various realistic traffic scenarios rather than its maximum achievable throughput.

To enable a fair comparison, all three traffic patterns are scaled to have the same average arrival rate of 0.5, which corresponds to a mean inter-arrival time of two seconds. The traffic patterns in each scenario have been implemented correctly, as the kernel density estimation (KDE) closely matches the theoretical probability density function. KDE is a method of estimating the probability density function of a random variable from a finite sample of data. KDE is essentially a smoother version of a histogram, which makes comparison with the theoretical probability density function easier. Smooth curves can be used to visualize the distribution of a random variable. These curves, which represent the distribution of continuous variables, are called probability density functions (PDFs).

The empirical statistics for the traffic pattern with steady load are: 
\begin{itemize}
    \item \textbf{Empirical mean}: $\approx 1.93 s$
    \item \textbf{Empirical variance}: $\approx 3.59 s^2$
\end{itemize}

The theoretical statistics for the traffic pattern with steady load are:
\begin{itemize}
    \item \textbf{Theoretical mean}: $$ \frac{1}{\lambda} = \frac{1}{0.5} = 2s$$ 
    \item \textbf{Theoretical variance}: $$ \frac{1}{\lambda^2} = \frac{1}{0.5^2} = 4 s^2$$
\end{itemize}

\begin{figure}[H]
    \centering
    \includegraphics[width=0.9\linewidth]{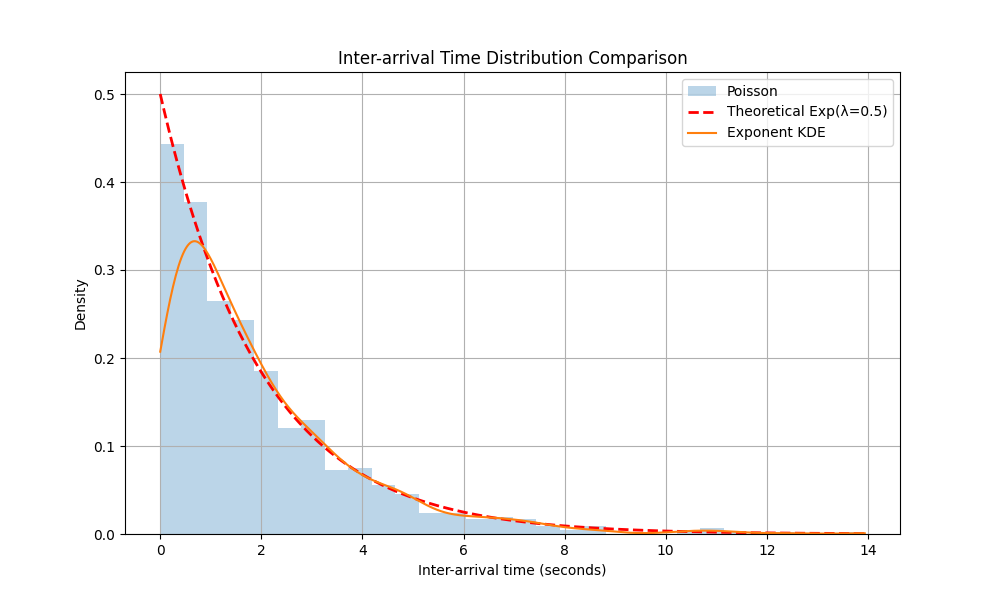}
    \caption{Empirical inter-arrival time distribution (histogram and KDE) compared with the theoretical Exponential distribution probability density function for a steady traffic pattern ($\lambda$ = 0.5)}
    \label{fig:steady_load}
\end{figure}

The empirical statistics for the traffic pattern with moderate burstiness ($\alpha$ = 1.2) are:
\begin{itemize}
    \item \textbf{Empirical mean}: $\approx 1.97 s$
    \item \textbf{Empirical variance}: $\approx 2.84 s^2$
\end{itemize}

The theoretical statistics for the Gamma distribution ($\alpha$ = 1.2), scaled to a mean of 2s, are:
\begin{itemize}
    \item \textbf{Theoretical mean}: 2s
    \item \textbf{Scale parameter $\theta$}: $$\theta = \frac{mean}{\alpha} = \frac{2}{1.2} \approx 1.67 $$
    \item \textbf{Theoretical variance}: $$\alpha*\theta^2 = 1.2*(1.67)^2 \approx 3.33 s^2$$
\end{itemize}

\begin{figure}[H]
    \centering
    \includegraphics[width=\linewidth]{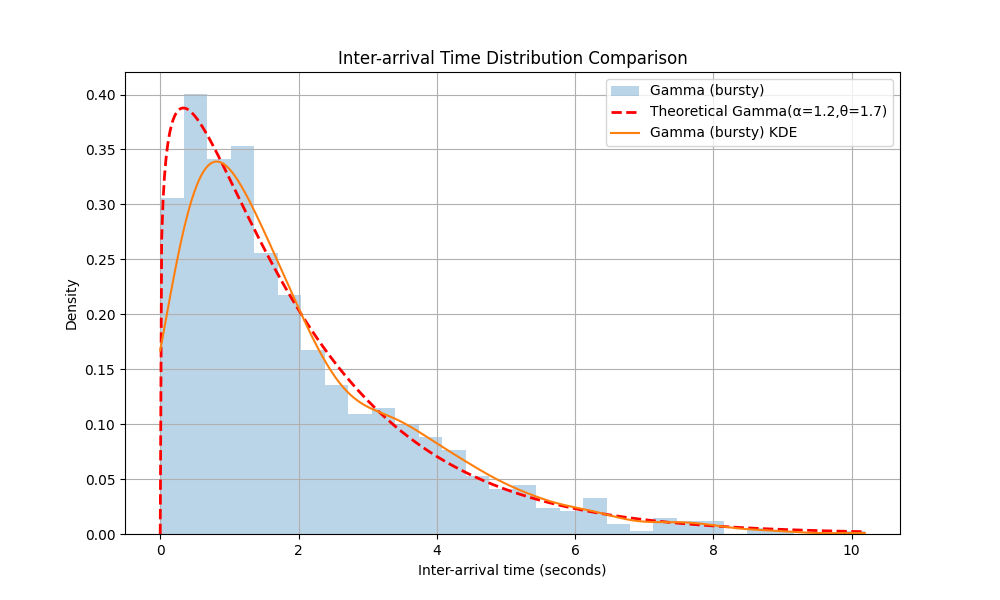}
    \caption{Empirical inter-arrival time distribution (histogram and KDE) compared with the theoretical Gamma distribution probability density function for a traffic pattern with moderate burstiness (shape parameter $\alpha$ = 1.2)}
    \label{fig:moderate_load}
\end{figure}

The empirical statistics for traffic pattern with extreme burstiness ($\alpha$ = 0.8) are
\begin{itemize}
    \item \textbf{Empirical mean}: $\approx 1.95 s$
    \item \textbf{Empirical variance}: $\approx 4.64 s^2$
\end{itemize}

The theoretical statistics for the Gamma distribution ($\alpha$ = 0.8), scaled to a mean of 2s, are:
\begin{itemize}
    \item \textbf{Theoretical mean}: $2 s$
    \item \textbf{Scale $\theta$}: $$\theta = \frac{mean}{\alpha} = \frac{2}{0.8} \approx 2.5 $$
    \item \textbf{Theoretical variance}: $$\alpha*\theta^2 = 0.8*2.5^2 = 5 s^2$$
\end{itemize}

\begin{figure}[H]
    \centering
    \includegraphics[width=\linewidth]{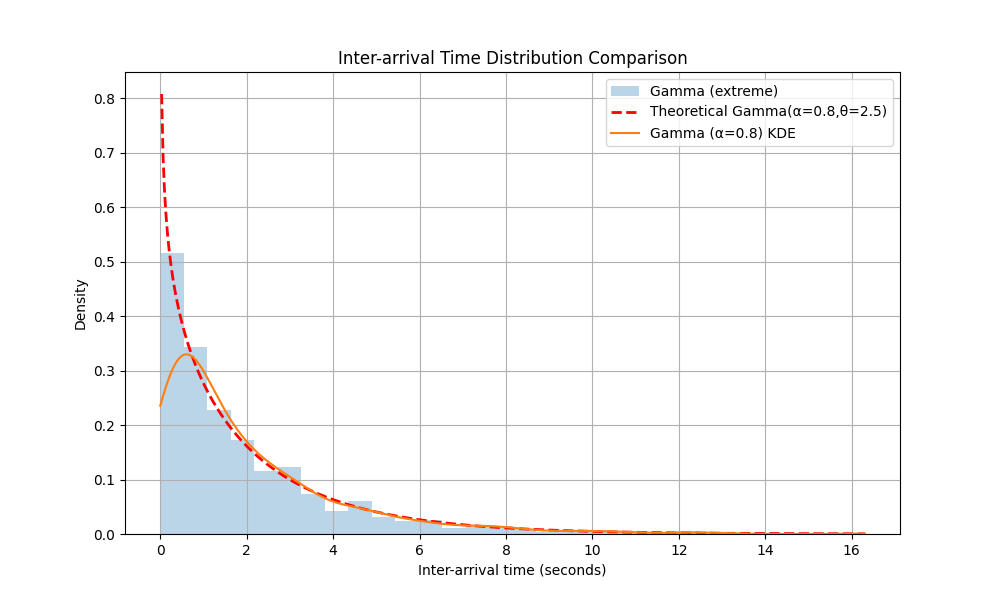}
    \caption{Empirical inter-arrival time distribution (histogram and KDE) compared with the theoretical Gamma distribution probability density function for a traffic pattern with extreme burstiness (shape parameter $\alpha$ = 0.8)}
    \label{fig:extreme_load}
\end{figure}

Overall, Figures \ref{fig:steady_load}, \ref{fig:moderate_load}, and \ref{fig:extreme_load} demonstrate that the kernel density estimate (solid orange line) closely matches the theoretical probability density function (dashed red line). Together with the comparison of empirical and theoretical values, this indicates that all three traffic patterns have been implemented correctly.

\section{Experimental Setup}

\subsection{Infrastructure}
This study involved optimizing, evaluating, and testing on a server with two Intel Xeon Gold 6230 CPUs at 2.10 GHz, eight Quadro RTX 8000s with 48 GB of VRAM each, and 693 GB of RAM. Although the server has eight GPUs, only one was used for these tasks. The system utilized CUDA v13.0, BentoML v1.4.35, ONNX v1.20.1, PyTorch v2.9.1+cu128, and Locust v2.43.3. 

\subsection{Model Selection}
This study uses the pretrained RoBERTa Base model provided by graphworks.ai\footnote{https://graphworks.ai}, which was fine-tuned for sentiment analysis. This FP32 PyTorch model serves as the reference in the evaluation and load testing, along with the use of the simpletransformers. 
\subsection{Dataset Selection}
The \textbf{Sp1786/Multiclass-Sentiment-Analysis-Dataset}\footnote{https://huggingface.co/datasets/Sp1786/multiclass-sentiment-analysis-dataset}, which includes positive, negative, and neutral categories, is used to perform load tests and evaluate the performance of different system variants. To ensure a fair comparison, the same subset is applied in different experiments, using the same random seed and fixed number of 1,000 examples. 

\section{Results and Discussion}
This section presents the results of the experimental evaluation. It also analyzes improvements in performance and observed trade-offs in the system. Furthermore, it discusses the implications for scalable AI inference using BentoML-based predictors.
\subsection{Model Optimization}
This section discusses the results of evaluating different approaches to optimizing the model. In addition to the baseline implementation using simpletransformers with FP32 PyTorch, the various optimized approaches are evaluated across an inference data batch size ranging from 1 to 32. The key findings are presented below.
\begin{itemize}
    \item \textbf{Model Size:} In this study, the FP32 models used for evaluation, which are either PyTorch or ONNX, require approximately 498.7 MB of storage. However, an exact 2x reduction in memory footprint is achieved when the model size is reduced to 249.4 MB by converting the model into FP16. This reduction is the result of the half-precision representation. Instead of using 32 bits for each weight, only 16 bits are used.
    \item \textbf{Latency:} As Table \ref{tab:model_evaluation_latency} shows, the baseline implementation reveals a latency of 2.18 seconds per sample at a batch size of 1 and 0.14 seconds per sample at a batch size of 32. In contrast, all the optimized variants achieve latency in the millisecond to sub-millisecond range per sample. Among these, FP16 ONNX delivers the lowest per-sample latency across all tested batch sizes. This significant improvement makes FP16 ONNX suitable for scenarios requiring low individual prediction latency, such as real-time and online applications.
    \item \textbf{Throughput:} Table \ref{tab:model_evaluation_throughput} shows that the baseline implementation achieves a maximum throughput of 7.37 samples per second with a batch size of 32. This value is too low compared to the optimized variants, which can scale to hundreds or thousands of samples per second. Half-precision inference achieves the greatest improvement, as FP16 PyTorch and FP16 ONNX peak at nearly 2,000 samples per second. These results demonstrate the effectiveness of combining the ONNX format with the lower-precision FP16 format for high-throughput and batch-processing workloads.
    \item \textbf{Accuracy \& F1 Scores:} To evaluate generalization to unseen data, we tested the trained sentiment analysis model on a fixed subset of 1,000 samples from an external Hugging Face dataset. The dataset has a relatively balanced class distribution, with 285 negative, 323 positive, and 392 neutral instances.
    
    All evaluated variants achieved identical results: an accuracy of 0.42 (42\%), a Macro F1 score of 0.32, and a Weighted F1 score of 0.34. This shows that exporting the model to ONNX, applying graph optimizations, and using FP16 do not reduce classification performance.
    
    Since the external dataset uses the same three sentiment labels --- positive, negative, and neutral --- the low scores cannot be attributed to a label mismatch. A more plausible explanation is domain shift. The model was trained on news articles in the energy domain, whereas the external evaluation dataset contains short, informal texts from a different source. Differences in style, vocabulary, text length, and sentiment expression likely contributed to the lower performance.
    
    The slightly higher Weighted F1 compared with Macro F1 suggests somewhat better performance on more frequent classes. Overall, although the accuracy and F1 scores are lower than expected for in-domain sentiment analysis, they do not indicate that the model itself is ineffective. Rather, they reflect the challenge of applying the model to data from a different domain. Importantly, all optimization variants preserved the same predictive quality, meaning that the improvements in latency, throughput, and model size were achieved without additional accuracy loss. Further improvement would likely require domain adaptation or retraining on data closer to the target domain.

\end{itemize}

Overall, these results highlight the effectiveness and performance improvements, as well as the storage savings, when the AI model is exported to the ONNX format with half-precision computation, graph optimization, and runtime-specific optimization. FP16 ONNX requires half the disk space and provides the best performance. For production environments, where storage, memory bandwidth, and inference speed play an important role, the ONNX format with lower-precision FP16 is the best option. Additionally, these findings highlight the importance of runtime selection and precision tuning when optimizing a model for inference, as well as the trade-off between performance and convenience. The simpletransformers used in the baseline implementation is a high-level wrapper built on top of Hugging Face Transformers, which explains this trade-off. While the simpletransformers can be a good choice for prototyping and training purposes, it is not suitable for production environments.

\begin{table}[H]
\caption{Latency (ms) - Lower is better}
\vspace{3mm}
\label{tab:model_evaluation_latency}
\centering
\small
\begin{tabular}{lcccccc}
\toprule
 & \multicolumn{6}{c}{\textbf{Batch Size}} \\
\cmidrule{2-7}
\textbf{Evaluation Type} & \textbf{1} & \textbf{2} & \textbf{4} & \textbf{8} & \textbf{16} & \textbf{32} \\
\midrule
Base (FP32 PyTorch) & 2177.49 & 1197.24 & 727.27 & 500.12 & 266.99 & 135.72 \\
Opt. (FP32 PyTorch) & 6.38 & 3.40 & 2.61 & 2.41 & 2.22 & 2.09 \\
Opt. (FP16 PyTorch) & 6.22 & 3.03 & 1.53 & 0.79 & 0.6 & 0.56 \\
Opt. (FP32 ONNX)    & 4.09 & 2.97 & 2.69 & 2.48 & 2.49 & 2.43 \\
Opt. (Opt. ONNX)   & 3.89 & 2.73 & 2.25 & 2.1 & 2.09 & 2.05 \\
Opt. (FP16 ONNX)  & 1.95 & 1.12 & 0.76 & 0.57 & 0.53 & 0.53 \\
\bottomrule
\end{tabular}
\end{table}

\begin{table}[H]
\caption{Throughput (samples/s) - Higher is better}
\vspace{3mm}
\label{tab:model_evaluation_throughput}
\centering
\small
\begin{tabular}{lcccccc}
\toprule
 & \multicolumn{6}{c}{\textbf{Batch Size}} \\
\cmidrule{2-7}
\textbf{Evaluation Type} & \textbf{1} & \textbf{2} & \textbf{4} & \textbf{8} & \textbf{16} & \textbf{32} \\
\midrule
Base (FP32 PyTorch) & 0.46 & 0.84 & 1.38 & 2 & 3.75 & 7.37 \\
Opt. (FP32 PyTorch) & 156.85 & 293.81 & 383.77 & 415.52 & 450.09 & 478.01 \\
Opt. (FP16 PyTorch) & 160.71 & 329.66 & 652.07 & 1263.23 & 1666.86 & 1774.72 \\
Opt. (FP32 ONNX)    & 244.51 & 336.54 & 372.01 & 403.7 & 402.05 & 412.14 \\
Opt. (Opt. ONNX)    & 256.91 & 366.47 & 443.86 & 476.14 & 478.18 & 488.9 \\
Opt. (FP16 ONNX)    & 512.94 & 891.08 & 1309.46 & 1758.22 & 1905.78 & 1901.56 \\
\bottomrule
\end{tabular}
\end{table}

\subsection{Load Testing}
This section analyzes the results of load testing to evaluate the system’s performance under different traffic conditions. Key Findings:
\begin{itemize}
    \item \textbf{Failure Rate:} In all three scenarios, the optimized and baseline approaches operate with perfect reliability (0\% error). 
    \item \textbf{Latency:} Table \ref{table:load_testing_latency} presents the latency across three scenarios. The baseline approach exhibits extremely high latency across three scenarios. This level of delay negatively impacts user experience in real-world scenarios and renders it practically unsuitable for real-time or interactive applications, even under moderate or extreme request rates. In contrast, optimized approaches have a latency that is a hundred times lower. 
    \item \textbf{Throughput:} The optimized approaches can process nearly 0.5 requests per second, whereas the baseline approach can only process 0.2.
    \item \textbf{Response Time:} As Table \ref{table:load_testing_response_time} shows, the baseline approach has poor performance and high response times under different loads. These values make the baseline approach unsuitable for latency-sensitive or production-like workloads. In contrast, dramatic improvements in response time can be observed in all the optimized approaches. 
    \item \textbf{Total Load Test Duration:} Due to high latency and low throughput, the baseline approach takes over 80 minutes to complete the load test in different scenarios. In contrast, all of the optimized variants complete the load test in approximately 30 minutes.
\end{itemize}

Overall, the dramatic improvements brought by all the optimized variants are clearly visible in the three load tests with different scenarios, compared to the baseline approach.
\begin{itemize}
    \item \textbf{Throughput:} In all three test scenarios, the throughput of the optimized variants was more than twice that of the baseline approach. This number was even achieved by the FP32 PyTorch variant. The baseline approach also uses FP32 PyTorch. However, the difference is that the baseline approach uses a simpletransformer. While this library can simplify implementation by abstracting most processes, it can also complicate optimization and introduce unnecessary overhead, impacting performance. Instead of using the simpletransformers library, all of the optimized variants use the Hugging Face Transformers library directly. This library requires manual implementation of processes such as prediction and tokenization. Additionally, optimizations are applied at different levels, such as the model and service levels. Consequently, the RPS improves under realistic scenarios.
    \item \textbf{Latency:} Similar to throughput, clear improvements in latency are observed. Compared to the baseline approach, the optimized variants have latencies that are tens to hundreds of times lower. Among the optimized variants, the ONNX-based variants deliver the lowest median and near-worst-case latencies consistently. Besides directly adapting the Hugging Face Transformers library, applying ONNX export, graph optimization, and half-precision computation transforms the baseline approach, which has extremely poor performance, into a highly responsive inference service. These variants are ideal for deployment scenarios requiring low latency and robust tail performance.
    \item \textbf{Response Time:} Exporting the model to ONNX and optimizing the graph provides massive speedups and greatly improves stability and predictability under realistic request patterns. Using half-precision inference in most cases further improves these benefits.
\end{itemize}

\begin{table}[H]
\centering
\caption{Latency (ms) comparison across scenarios - Lower is better}
\label{table:load_testing_latency}
\vspace{3mm}
\begin{tabular}{l|ccc|ccc|ccc}
\hline
 & \multicolumn{3}{c|}{\textbf{Scenario 1}} 
 & \multicolumn{3}{c|}{\textbf{Scenario 2}} 
 & \multicolumn{3}{c}{\textbf{Scenario 3}} \\
\textbf{Evaluation Type}
 & p50 & p95 & p99 
 & p50 & p95 & p99 
 & p50 & p95 & p99 \\
\hline
Base (FP32 PyTorch)  & 2700 & 4100 & 4200 & 3000 & 4100 & 4200 & 3100 & 4100 & 4200 \\
Opt. (FP32 PyTorch)  & 46 & 61 & 110 & 45 & 56 & 73 & 44 & 55 & 73 \\
Opt. (FP16 PyTorch)  & 46 & 57 & 92  & 45 & 59 & 85 & 44 & 55 & 71 \\
Opt. (FP32 ONNX)    & 27 & 36 & 56  & 26 & 36 & 55 & 26 & 36 & 53 \\
Opt. (Opt. ONNX)     & 26 & 38 & 59  & 26 & 35 & 53 & 26 & 35 & 52 \\
Opt. (FP16 ONNX)     & 27 & 40 & 65  & 26 & 35 & 55 & 26 & 35 & 54 \\
\hline
\end{tabular}
\end{table}

\begin{table}[H]
\centering
\caption{Response Time (ms) comparison across scenarios - Lower is better}
\label{table:load_testing_response_time}
\vspace{3mm}
\begin{tabular}{l|ccc|ccc|ccc}
\hline
 & \multicolumn{3}{c|}{\textbf{Scenario 1}} 
 & \multicolumn{3}{c|}{\textbf{Scenario 2}} 
 & \multicolumn{3}{c}{\textbf{Scenario 3}} \\
\textbf{Evaluation Type}
 & Min & Avg & Max 
 & Min & Avg & Max 
 & Min & Avg & Max \\
\hline
Base (FP32 PyTorch)  & 2169.75 & 2844.81 & 4287.7 & 2337 & 2931.33 & 4287.49 & 2329.78 & 3015 & 4343.67 \\
Opt. (FP32 PyTorch)  & 17.41 & 48.62 & 346.6 & 19.31 & 45.95 & 135 & 18.53 & 44.8 & 154.38 \\
Opt. (FP16 PyTorch)  & 21.56 & 47.65 & 230  & 16.45 & 46 & 428 & 22.44 & 44.56 & 128.3 \\
Opt. (FP32 ONNX)     & 18.31 & 28.83 & 213.78 & 12.22 & 27.84 & 493.66 & 13 & 27.47 & 130.52 \\
Opt. (Opt. ONNX)     & 13.27 & 27.92 & 108.56  & 13.18 & 27.3 & 91.88 & 13.8 & 27.8 & 951.49 \\
Opt. (FP16 ONNX)     & 15.5 & 28.47 & 243 & 12.73 & 27.48 & 570 & 15.42 & 27 & 483.5 \\
\hline
\end{tabular}
\end{table}

\subsection{Resilience Testing}
This section analyzes the results of the resilience test, which evaluate the system’s resilience and the effectiveness of the automated service recovery mechanisms.
\begin{figure}[H]
    \centering
    \includegraphics[width=\linewidth]{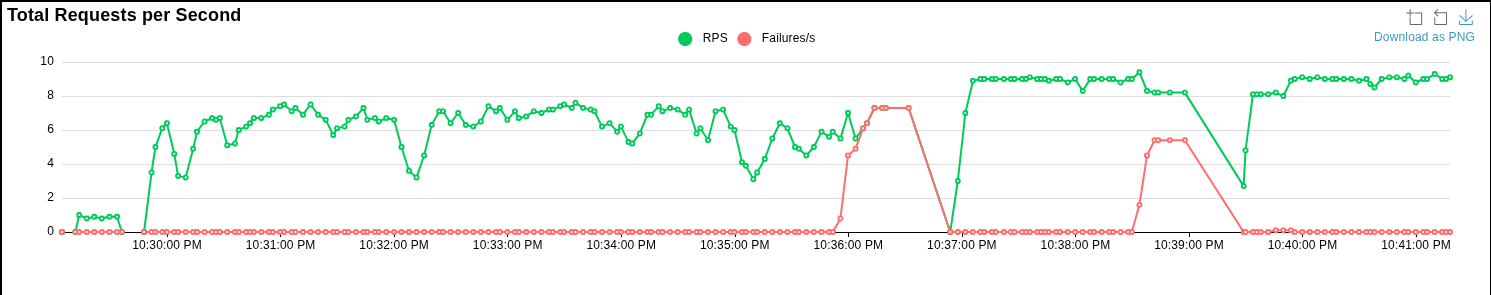}
    \caption{Service Behavior under the Resilience Test}
    \label{fig:resilience_testing}
\end{figure}

Figure \ref{fig:resilience_testing} illustrates the behavior of a single-node K3s cluster during the resilience test. Additionally, during the resilience test, only one type of error was observed. The POST method /analyze-sentiment produced 216 occurrences of CatchResponseError ("Failed: 0"), indicating a recurring failure pattern under test conditions. 

There is no increase in error rates when the Kubernetes Control Plane and containerd are restarted. This may be because the restart process only lasts a very short time. Although the RPS does not drop to 0, it still fluctuates, as shown by the RPS graph. A rapid rise in error rates occurs when the Deployment restarts for the first time. At the same time, growth in response time is also visible. This occurs because when the Deployment restarts, the old ReplicaSet and Pod are replaced by a new one. It takes time for the new Pod to become ready. At this point, the traffic stabilizes. 

A second drop in RPS and a second increase in error rates occur when the current Pod is terminated and replaced by a newly created Pod. Once the application is fully operational, the number of failures drops to zero, and RPS stabilizes. The experimental results demonstrate that the system can self-heal, even in a single-node cluster.

When a disruption occurs under continuous load — such as the termination of a Pod — Kubernetes can detect the failure and automatically recreate the Pod. The service is restored without any manual intervention. Despite the service disruption during container restarts, the application recovers and stabilizes autonomously. Compared to the current Docker-only deployment, which requires manual intervention, this demonstrates clear improvements in resilience.

\subsection{Practical Implications}
These findings have practical implications for using the BentoML-based predictor to perform scalable AI inference in production environments.
\begin{itemize}
    \item \textbf{Model Optimization Trade-offs:} The experimental results demonstrate improvements in various metrics, including model size, latency, and throughput. Despite the reduction in numerical precision from FP32 to FP16, the predictive performance of the evaluated types remains unaffected. The accuracy and F1 scores are identical. Meanwhile, the FP16 variant reduces model size and inference latency. Performance improves further when the ONNX format and FP16 are combined. 
    
    Reducing the memory footprint and achieving faster inference times enables higher throughput and improved scalability. This is useful when the system is subject to concurrent workloads in a production environment. However, this observation is specific to the evaluated model architecture and task. Often, when other models or workloads are applied, the system experiences a reduction in accuracy as precision decreases. This emphasizes the importance of evaluating and validating models when applying lower-precision optimization to different contexts.
    \item \textbf{Configuration Recommendations for BentoML Predictors:} Evaluation and testing show that the performance of the BentoML predictor is highly sensitive to configuration parameters, such as adaptive batching. Enabling adaptive batching improves throughput under moderate to high traffic conditions. Based on these results, adaptive batching is recommended for most production deployments, with further tuning based on workload characteristics.
    \item \textbf{Workload-aware Deployment Strategies:} The load testing experiments used exponentially distributed request arrivals to model steady traffic conditions and gamma-distributed interarrival times to represent varying degrees of traffic burstiness. These models allowed for the systematic evaluation of the predictor’s behavior under stable and highly variable workloads. Higher burstiness in the traffic could create load spikes that impact performance metrics such as latency and throughput. Adapting the configuration — such as implementing adaptive batching — improves and stabilizes service performance. These findings demonstrate the importance of deploying strategies informed by expected arrival patterns.
    \item \textbf{Resilience Considerations:} The resilience evaluation discusses the system’s ability to recover from runtime disruptions without human intervention. The focus is on resilience rather than high availability. Since the baseline deployment consists of only a standalone Docker-based service, its resilience is limited. When failures or disruptions occur, manual restarts and intervention are required to restore service availability.

    In contrast, the optimized deployment with K3s showed improved resilience, even when the service was deployed on a single-node cluster with a single service replica. Despite the disruptions we simulated during the experiment, the service automatically restored itself to a functional state thanks to Kubernetes’s built-in mechanisms. These results show that container orchestration can significantly improve service resilience, even when deployed on minimally configured clusters.

    Regarding deployment in a production environment, these results suggest that migrating from a standalone container deployment to a lightweight Kubernetes distribution, such as K3s, can provide resilience benefits. However, adapting additional replicas or a multi-node configuration should be considered in the future. This adaptation will ensure higher availability.

\end{itemize}
\section{Summary and Conclusion}
\subsection{Limitations}
\begin{itemize}
    \item \textbf{Dataset Constraints:} The limited number of datasets may affect the evaluation results. For future work, it is important to consider more diverse and larger datasets. 
    \item \textbf{Model Quantization:} Reducing the number of bits offers many advantages, such as requiring less memory storage and improving performance. Although the experiment showed an improvement in performance when the model was converted to use lower-precision FP16, future work could consider using even lower precision, such as FP8, INT8, or BF16, depending on what the hardware supports.
    \item \textbf{Scalability and Reliability via Kubernetes:} A single-node K3s cluster cannot provide high availability, but it does provide automated recovery. Future work should extend this setup to a multi-node cluster with multiple replicas to achieve fault tolerance and near-zero downtime. 
    \item \textbf{Concurrent Request Handling:} Workers are processes that respond to code logic execution within a BentoML service. The BentoML-based AI inference system by graphworks.ai serves multiple machine learning (ML) models for various tasks on a system with multiple graphics processing units (GPUs). Assigning specific GPUs to different workers allows each worker to process tasks independently. This can lower total inference time, boost throughput, and optimize parallel processing.
\end{itemize}
\subsection{Conclusion and Outlook}
This study evaluated the performance of the AI predictor across different areas. Following evaluation and testing, potential optimizations were suggested for a scalable, BentoML-based AI inference predictor. These optimizations focus on the model, service, and deployment levels. The study examined how model precision, system configuration, and different workload patterns affect inference performance, scalability, and resilience through experimentation and evaluation.

The results demonstrate that optimizing the model significantly improves inference efficiency without compromising predictive performance. Specifically, reducing numerical precision from FP32 to FP16 improves performance in several areas, including higher throughput, lower latency, and a smaller model size. This optimization maintains the same accuracy and F1 scores for the evaluated model and dataset. Furthermore, using the ONNX format with FP16 improves performance even more. These results imply that, when supported by the underlying hardware and inference framework, using the ONNX format with FP16 is an effective and practical optimization strategy in production environments.

Load testing under stochastic traffic patterns revealed the system's behavior under various realistic operating conditions. Steady traffic was modeled as a Poisson process. Thus, the inter-arrival times follow an Exponential distribution. Other traffic patterns with varying degrees of burstiness were modeled using the Gamma distribution. The experiments revealed that system performance is sensitive to workload characteristics.

Resilience experiments have demonstrated that even a minimal K3s deployment with a single replica can automatically recover from service disruptions. This finding represents a practical improvement over standalone container-based deployments because it increases system resilience with container orchestration.

In summary, this study shows that BentoML can provide a solid foundation for scalable AI inference when optimization and configuration choices are based on performance evaluations. The findings offer practical suggestions for selecting model precision, tuning service parameters, and deploying inference workloads under different traffic conditions. Future research could build on this evaluation by exploring additional optimization techniques, hardware accelerators, and large-scale deployment scenarios.

%
%
%
\bibliographystyle{splncs04}
\bibliography{main}

\end{document}